# How Transferable are CNN-based Features for Age and Gender Classification?

Gökhan Özbulak[1], Yusuf Aytar[2] and Hazım Kemal Ekenel[1]

**Abstract:** Age and gender are complementary soft biometric traits for face recognition. Successful estimation of age and gender from facial images taken under real-world conditions can contribute improving the identification results in the wild. In this study, in order to achieve robust age and gender classification in the wild, we have benefited from Deep Convolutional Neural Networks based representation. We have explored transferability of existing deep convolutional neural network (CNN) models for age and gender classification. The generic AlexNet-like architecture and domain specific VGG-Face CNN model are employed and fine-tuned with the Adience dataset prepared for age and gender classification in uncontrolled environments. In addition, task specific GilNet CNN model has also been utilized and used as a baseline method in order to compare with transferred models. Experimental results show that both transferred deep CNN models outperform the GilNet CNN model, which is the state-of-the-art age and gender classification approach on the Adience dataset, by an absolute increase of 7% and 4.5% in accuracy, respectively. This outcome indicates that transferring a deep CNN model can provide better classification performance than a task specific CNN model, which has a limited number of layers and trained from scratch using a limited amount of data as in the case of GilNet. Domain specific VGG-Face CNN model has been found to be more useful and provided better performance for both age and gender classification tasks, when compared with generic AlexNet-like model, which shows that transfering from a closer domain is more useful.

**Keywords:** Soft biometrics, age classification, gender classification, convolutional neural networks, deep learning, transfer learning.

## 1 Introduction

The effectiveness of soft biometric traits was argued in [JDN04a] and proven to be applicable on primary biometric applications such as face and fingerprint recognition in several studies [JDN04b,NPJ10,De12]. Soft biometric traits such as age, gender, height, and weight were used in these studies not as the identity indicators but as the complementary information for primary biometric systems in a fusion framework. A recent survey on soft biometrics [DER15] highlights the benefits of using these kinds of traits in order to be more descriptive and representative while working with biometric data.

Automatic estimation of age and gender from facial images has received an increasing attention in recent years. In terms of soft biometrics applications, the developed systems are expected to process the real-world data obtained from uncontrolled environments, such as customs, airports, and stadiums; in varying conditions such as illumination changes or facial occlusion (*i.e.*, glass, beard). Recent advancements in deep convolutional neural networks, increasing computational capabilities, and publicly available high amount of data have made it possible to have robust classifiers for age and gender classification as for many other computer vision problems. Today, deep CNN-based features are preferred over hand-crafted features due to significant performance increase they have provided. However, for some computer vision tasks, large amount of annotated data may not be available for deep CNN training. Still, recent studies [Ra14, HL06, Yo14, Gi14, Oq14] have shown that a pretrained deep CNN model can be transferred and used for another image classification problem. The idea of transferring pretrained deep CNN models is also very practical, since training a deep CNN model from scratch requires significant computational resources and time.

---

[1] Dept. of Computer Engineering, Istanbul Technical University, Turkey, {gokhan.ozbulak, ekenel}@itu.edu.tr
[2] CSAIL, Massachusetts Institute of Technology, USA, yusuf@csail.mit.edu

In this paper, we have explored transferability of two pretrained deep CNN models, namely, AlexNet-like CNN model and VGG-Face CNN model [PVZ15] for age and gender classification. The former is a generic model trained for general visual recognition, whereas, the latter is a face-specific model trained for face recognition. We will refer to these models as generic and domain-specific, respectively. These transferred models are then compared to the state-of-the-art age and gender classification system, which is also based on a deep CNN model –GilNet [LH15]. However, this model is learned specifically for age and gender classification. We will refer to this model as task-specific model. The experiments have been conducted on the Adience dataset [EEH14], which contains face images collected under uncontrolled conditions.

The contribution of this paper can be summarized as follows: First, we have shown that generic and domain specific deep CNN models can be transferred successfully for age and gender classification problems. Second, by using appropriate transfer learning approaches, a pretrained CNN model can perform even better than training a new task specific CNN model from scratch. Using a transferred model, we have attained 7% and 4.5% absolute performance improvement over using a task-specific model for age and gender classification, respectively. Third, transferring a domain specific CNN model rather than a generic CNN model is found to be more useful, yielding better classification results, *i.e.*, 57.9% vs. 52.3% for age estimation and 92.0% vs. 90.5% for gender classification. Overall, we have shown that when limited amount of data is available for a specific computer vision task, it is better to benefit from a pretrained deep CNN model and transfer it to the task at hand, instead of building and training from scratch a task-specific CNN model with relatively less number of layers as in GilNet [LH15]. It has also been found that in terms of transferability, employing a pretrained deep CNN model, which is trained with the data from a closer domain, would be more beneficial.

## 2    Related work

In this section, previous work on age and gender classification is reviewed with the benchmarks prepared for these tasks.

### 2.1    Datasets

Age and gender benchmarks can be categorized as datasets taken from controlled or uncontrolled environments. FG-NET Aging [La15], MORPH [RT06], and UIUC-IFP-Y Internal Aging [FH08] are benchmarks that have facial images with gender and accurate age information. These datasets, however, are prepared by images taken under controlled conditions and do not reflect real-world situations. Similarly, FERET benchmark [Ph98] has facial images obtained in constrained environments with gender information.

PubFig [Ku09], Gallagher group photos [GC09], and Adience [EEH14] benchmarks have facial images with gender and age group information. The images in these datasets are taken in unconstrained environments and represent real-world challenges. Adience benchmark [EEH14] is stated to be more challenging and suitable for age and gender estimation when compared with other benchmarks according to [EEH14] (see Figure 1). Further information about the benchmarks explained above can be found in [EEH14].

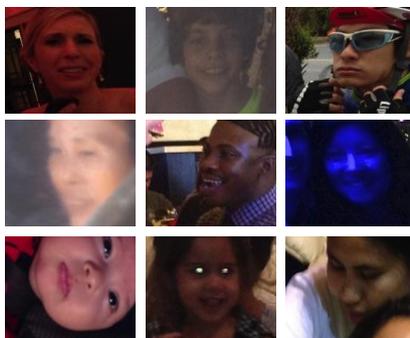

Fig. 1: Sample images from challenging Adience dataset

### 2.2    Age classification

Previous approaches for age estimation considered the ratio between facial features such as eyes, nose and mouth [KL94], manifold learning [Gu08], global features with Active Appearance Models (AAM) [LTC02], local features such as Gabor, Gabor+Local Binary Pattern (LBP) combined with linear classifiers [GA09, Ch11]. In a recent study [Po15]; AAM, LBP, Gabor features are integrated with Local Phase Quantization (LPQ) in order to obtain a feature representation and that representation is used for estimating age group first and then classifying the exact age with SVM and Support Vector Regression (SVR).

A state-of-the-art deep CNN approach for age classification, which is also used as the baseline method in this paper, is GilNet model [LH15]. This CNN has a shallow architecture when compared with complex CNNs such as AlexNet [KSH12] or VGG-Face [PVZ15] and was trained with Adience dataset [EEH14] from scratch. GilNet classifies the faces into eight different age groups rather than exact or apparent age.

A recent competition, named as ChaLearn Looking at People (LAP) [Es15] focuses on estimation of apparent age rather than chronological exact age. The successful approaches competed in this challenge are mostly based on deep CNN [RTG15, KCW15, Ra15] that shows the importance and impact of CNNs in age estimation, which in return motivates us to conduct this comprehensive and comparative study on the use of CNN-models for age and gender classification.

### 2.3    Gender classification

Gender estimation based on raw pixel intensities were used with classifiers such as SVM and AdaBoost in [MY02, DBD14, BR07]. The performance of [MY02] and [BR07] were reported on controlled FERET dataset [Ph98], whereas [DBD14] was experimented with some uncontrolled datasets including Labeled Faces in the Wild (LFW) [Hu07].

Deep CNN models have also been utilized for gender classification in recent years [LH15]. GilNet model, proposed in [LH15], is trained with Adience dataset [EEH14] from scratch for gender estimation. In [ABD16], a CNN ensemble model is proposed for running efficiently in embedded devices. Experiments of this method are conducted on LFW [Hu07] dataset and state-of-the-art results are obtained according to [ABD16]. The study in [MAP16] combines local features and deep networks by learning from local overlapping patches in feed-forward manner.

Different from these CNN-based approaches for age and gender classification, we, in this study, show the effectiveness of the transferability for domain-specific features, which are extracted from facial objects, over generic features, which are obtained from general objects, with limited data. We also show that transferring the models would work better than training a task-specific network from scratch (*i.e.,* GilNet [LH15]) for these estimation tasks.

## 3    Deep convolutional neural networks for age and gender classification

This section describes two transfer learning strategies (fine-tuning and CNN features with SVM classifiers) applied on pre-trained deep convolutional neural network models in order to construct robust classifiers for age and gender classification tasks. The topologies of these pre-trained models are also explained in detail.

### 3.1    Task specific GilNet CNN model

**Architecture.** GilNet CNN model has three convolutional layers followed by three fully connected layers. Each convolution layer is activated by a Rectified Linear Unit (ReLU) and then max-pooled. The output of the first two convolutional layers is normalized by Local Response Normalization (LRN) approach proposed in [KSH12]. First two fully connected layers, FC6 and FC7, have an output size of 512 and both vectors are regularized by dropout with probability of 0.5 in order to obtain sparse feature representations. The last fully connected layer (FC8) learns the classification for age and gender estimation tasks. The very last layer, softmax, has a number of outputs based on the task in interest (two for gender task, *i.e.,* male/female classes) and each output corresponds to a probability of assigning the input to that output/label. This architecture is used for both age and gender tasks by training corresponding subsets of the Adience dataset [EEH14].

Input to the GilNet model is three-channel, 256x256 resolution image. The input is randomly augmented with cropping patches of 227x227 and mirroring in order to increase the number of samples. For each iteration, 50 samples of inputs are fed into the network for training. For further details of GilNet architecture, see [LH15].

**SVM classification on CNN features.** In this study, CNN features of both age and gender GilNet models are extracted from second fully connected layer (FC7) and a 512 dimensional feature vector is obtained for each training image. All feature vectors are then used for training SVM classifiers in grid-search manner: Both linear and RBF kernels are combined with cost values of C={0.001, 0.01, 0.1, 1, 10, 100, 1000}, ending up 14 SVM models for training/testing.

### 3.2    Generic AlexNet-like CNN model

**Architecture.** In this study, the reference model shipped with Caffe deep learning framework [Ji14] is used as generic deep CNN. The architecture of this network is very similar to AlexNet [KSH12], therefore it is named as AlexNet-like CNN model in this paper. AlexNet-like CNN only differs from AlexNet [KSH12] in the order of pooling and normalization layers and the data augmentation strategy. AlexNet-like CNN model has five convolutional layers followed by three fully connected layers (FC6, FC7 and FC8). Each convolution layer is activated by a ReLU and then max-pooled followed by LRN (In AlexNet [KSH12], ReLU activated output of each convolutional layer is first normalized with LRN and then max-pooled). FC6 and FC7 layers have an output size of 4096 with a dropout ratio of 0.5. The last fully connected layer (FC8) is adjusted to estimate a 1000-class object recognition task on ImageNet ILSVRC [Ru15]. The very last layer, softmax, has a probabilistic output of 1000 for each object category. This architecture was trained on 1.2M images of ImageNet ILSVRC challenge dataset [Ru15]. The input to AlexNet-like CNN is three channels, 256x256 resolution image. The input is randomly cropped with patches of 227x227 pixels and augmented for sampling. A batch size of 256 is used for each iteration.

**Fine-tuning.** The fine-tuning strategy in this study is as follows: The learning rate in the last fully connected layer (FC8) is set to a higher value (x10) than the learning rates of other layers. The reason of assigning upper layers higher learning rates is due to their dominant role in classification compared to lower layers, which provide mainly low level filtering. The number of outputs in FC8 is also changed to the number of classes for the new task, so that the network adapts its output to the new classification task. Once the model update based on steps mentioned above is implemented, the training procedure begins separately based on training sets of Adience dataset [EEH14] for age and gender classification.

**SVM classification on CNN features.** As in GilNet, here, the second fully connected layer (FC7) is used for feature extraction. The feature vector, which has a size of 4096, obtained from each training image is then fed into SVM classifiers, whose kernels are linear and RBF with cost values used in the GilNet case.

### 3.3   Domain specific VGG-Face CNN model

**Architecture.** VGG-Face CNN model [PVZ15] is the same as the VGGNet architecture [SZ15], which has 16 layers and is designed for object recognition, and differs mainly in training phase. VGG-Face CNN model [PVZ15] was trained on 2.6M facial images of 2622 identities, whereas VGGNet model was built upon ImageNet ILSVRC challenge [Ru15] by training on 1.3M images of 1000 class objects.

In VGG-Face CNN model [PVZ15], there are five convolutional layer blocks. For the first two blocks, one convolution layer is followed by another one and the output of each layer is activated by a ReLU. At the end of the block, the output is max-pooled in order to reduce the size. For the remaining three blocks, one convolution layer is followed by another convolution layer that is then followed by another one. Each convolution layer in the block is again activated by a ReLU and the final output is max-pooled. Different from other CNN models considered in this study, none of the convolution layer outputs is normalized (*i.e.,* with LRN). The last convolution layer block is followed by three fully connected layers (FC6, FC7 and FC8). FC6 and FC7 have an output size of 4096 with a dropout ratio of 0.5. FC8 is responsible for classification and has an output size of 2622 that is the number of identities defined in the training set. VGG-Face CNN model [PVZ15] is fed with a three channel input image, whose size is 224x224 pixels. Cropping and flipping are also applied to the training images. A batch size of 64 is used for each forward pass.

**Fine-tuning.** The same fine-tuning strategy followed in generic AlexNet-like model is applied.

**SVM classification on CNN features.** As in GilNet and AlexNet-like models, here, the output of FC7 layer (4096 dimensional feature vector) is used as the CNN feature and is fed into SVM classifiers with the same grid-search manner applied for previous models.

# 4    Experimental results

In this section, the results of the experiments conducted with fine-tuned AlexNet-like and VGG-Face CNN [PVZ15] models are reported and compared with the GilNet CNN model [LH15]. SVM models trained with the CNN features of the pre-trained networks are also compared with each other in order to show the effectiveness of the transferring the features from one domain to the other for age and gender classification tasks.

## 4.1    Adience benchmark

Adience benchmark [EEH14] has about 26K facial images of 2284 identities with age and gender labels. This dataset is categorized into eight age groups as {[0, 2], [4, 6], [8, 13], [15, 20], [25, 32], [38, 43], [48, 53], [60, -]} and each image has a gender label as male or female. The images were gathered from Flickr social photo sharing platform and no manuel filtering was applied for the images in the dataset. Therefore, the images in the Adience collection [EEH14] reflect the real-world characteristics very well as can be seen in Figure 1. The evaluation of the methods in this study is based on Adience protocol [EEH14]: For both age and gender estimation, five-fold cross-validation is applied to the aligned images of the dataset. Each fold has a split of train, validation, and test sets with a ratio of ~67%, ~8% and ~25%, respectively. For instance, the distribution of 17393 images in fold 1 is as 11703, 1312, and 4378 for train, validation, and test sets, respectively. The alignment method is applied as proposed in [EEH14].

## 4.2    Experimental settings

Caffe deep learning framework [Ji14] is used for fine-tuning the pre-trained models and extracting CNN features for SVM traning. Pre-trained AlexNet-like and VGG-Face models were retrieved from the Caffe's model repository named as "Model Zoo" while GilNet model was trained from scratch with Adience dataset instead of using the model in "Model Zoo". As the network configuration of all three models, a momentum of 0.9, a weight decay of 0.0005, a gamma of 0.1 with an initial learning rate of 0.001 were used as default parameters. VGG-Face model [PVZ15] was fine-tuned in 30000 iterations, while AlexNet-like and GilNet models were fine-tuned/trained in 50000 iterations. In the end, there were two models (one for age and one for gender) for each pre-trained model with a total of 6 models. The resulting fine-tuned models are named as Ft-AlexNet-like and Ft-VGG-Face (the prefix "Ft" refers to the "Fine-tuning").

GilNet, Ft-AlexNet-like and Ft-VGG-Face models were also used for CNN feature extraction in order to train those features with SVM classifiers in grid-search manner. For grid-search of the SVM models, linear and Radial Basis Function (RBF) kernels were used with varying cost values ranging from $10^{-3}$ to $10^3$ with a factor of 10, C={0.001, 0.01, 0.1, 1, 10, 100, 1000}, ending up 14 different SVM classifiers per task for each pre-trained model.

### 4.3    Results

**Age classification.** Ft-VGG-Face model is the winner in both validation and test sets for age estimation. In terms of validation accuracy, it is ~4% better than AlexNet-like model and ~7% better than GilNet model. The second best model is fine-tuned AlexNet-like model and it is ~3% better than GilNet. In Figure 2a, a comparison of the models over mean accuracy of five folds can be seen in detail (Note that each point in Figure 2a corresponds to average accuracy across the five folds for given iteration). An important point in the Figure 2a is that Ft-VGG-Face model can learn rapidly by reaching an accuracy over ~80% in just 1000 iterations (~4 epoch). This is very fast when compared with other models.

When the CNN features of these models are combined with SVM, Ft-VGG-Face model exhibits the best performance in both validation and test sets. On validation set, using feature representation of Ft-VGG-Face for training RBF kernel SVM with a cost of 100 has the best performance across SVM models as seen in Figure 2b. Ft-VGG-Face with RBF SVM of C=100 is ~12% better than the best GilNet+SVM setting and ~6% better than the best Ft-AlexNet-like+SVM model. In test set, Ft-VGG-Face+SVM setting is ~6% better than Ft-AlexNet-like+SVM model and ~9% better than GilNet+SVM model.

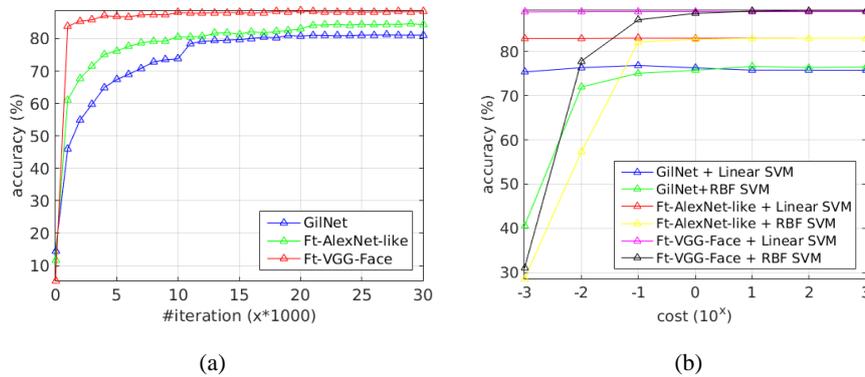

(a)                                    (b)

Fig. 2: The mean accuracy of transferred age models over five folds of train/validation set

As seen from Table 1, based on their best settings as reported on validation set, Ft-VGG-Face with RBF kernel of C=100 is the overall winner with a mean accuracy of 57.9%. This score is followed by Ft-VGG-Face with a mean accuracy of 57.2% and the third best model is Ft-AlexNet-like with a mean accuracy of 52.3%. The winner model has a significant performance gap when compared with GilNet (~7%) and the best method reported in [EEH14], which uses Local Binary Patterns and Four Patch LBP (FPLBP) as facial representation combined with dropout-SVM of 0.8, (~13%).

**Gender classification.** As in age classification, Ft-VGG-Face model is again the winner of the gender estimation task with a better validation accuracy of 3% than GilNet and ~1.5% than AlexNet-like models. Ft-AlexNet-like model is again the second best classifier for this task with a better accuracy of 1.5% than GilNet. The fold performance of the models in validation set can be seen in Figure 3a. As in age classification, here, Ft-VGG-Face model learns very fast by having the accuracies over ~95% in just 1000 iterations. As can be seen in Table 2, in test set, the similar performance holds: Ft-VGG-Face model has a mean accuracy of 91.9% followed by AlexNet-like model with a mean accuracy of 90.5%. GilNet has a mean accuracy of 87.5% as the worst CNN model.

| Models | F1 | F2 | F3 | F4 | F5 | Avg. |
|---|---|---|---|---|---|---|
| GilNet [LH15] | 56.9 | 44.7 | 56.6 | 46.7 | 48.8 | 50.7 |
| Ft-AlexNet-like | 59.8 | 46.6 | 56.5 | 47.2 | 51.7 | 52.3 |
| Ft-VGG-Face | 65.4 | 52.8 | 59.1 | 49.9 | 59.0 | 57.2 |
| GilNet+SVM (Linear,C=0.1) | 55.7 | 42.0 | 53.8 | 44.9 | 45.7 | 48.4 |
| Ft-AlexNet-like+SVM (Linear,C=0.1) | 58.7 | 44.1 | 55.5 | 47.6 | 50.4 | 51.2 |
| Ft-VGG-Face+SVM (RBF,C=100) | 66.6 | 53.4 | 58.7 | 52.7 | 58.4 | **57.9** |
| LBP+FPLBP+Dropout 0.8 (Best in [EEH14]) | - | - | - | - | - | 45.1 |

Tab. 1: Test results of age estimators in five folds with mean (as accuracy percentage)

In validation set, as seen in Figure 3b, the feature representation of Ft-VGG-Face with RBF SVM of C=100 has better performance of ~4% than the best GilNet+SVM setting (RBF kernel, C=100) and it is ~2% better than the best Ft-AlexNet+SVM setting (Linear kernel, C=1). The testing performance of Ft-VGG-Face combined with SVM is the best among the SVM based models with a mean accuracy of 92.0%. This is followed by Ft-AlexNet-like+SVM with a mean accuracy of 89.6% and then by GilNet+SVM model with a mean accuracy of 86.6%. Test results on folds can be examined in Table 2.

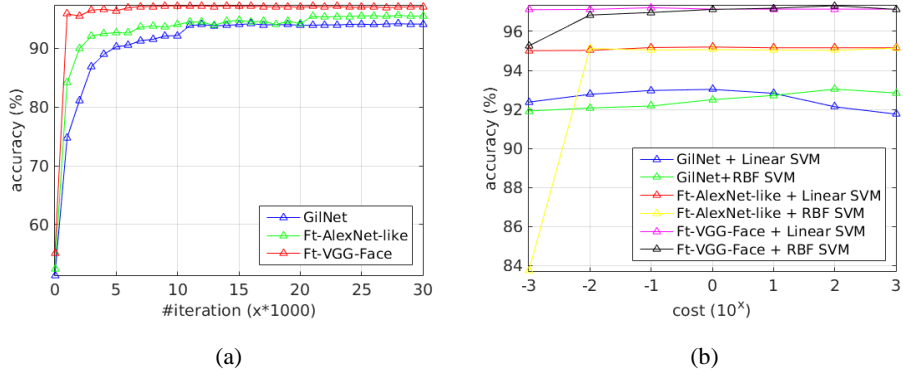

(a)                                        (b)

Fig. 3: The mean accuracy of transferred gender models over five folds of train/validation set

The overall winner on gender classification task is again Ft-VGG-Face with RBF kernel of C=100. As can be seen in Table 2, it has a mean accuracy of 92.0%. The second best model, Ft-VGG-Face achieves 91.9%, and 1.4% better than the third best model, Ft-AlexNet-like. The best performance of GilNet is 87.5%, which is worse by 4.5%, and the best approach in [EEH14], which is LBP and FPLBP combined with dropout-SVM of 0.5, has a mean accuracy of 77.8% that is ~14% worse than the Ft-VGG-Face models.

| Models | F1 | F2 | F3 | F4 | F5 | Avg. |
|---|---|---|---|---|---|---|
| GilNet [LH15] | 89.0 | 86.3 | 86.1 | 89.6 | 86.6 | 87.5 |
| Ft-AlexNet-like | 91.2 | 90.0 | 89.7 | 92.4 | 89.5 | 90.5 |
| Ft-VGG-Face | 93.3 | 92.6 | 89.4 | 94.4 | 89.9 | 91.9 |
| GilNet+SVM (RBF,C=100) | 88.9 | 85.9 | 85.6 | 87.9 | 85.1 | 86.6 |
| Ft-AlexNet-like+SVM (Linear,C=1) | 90.4 | 89.2 | 88.8 | 90.9 | 88.7 | 89.6 |
| Ft-VGG-Face+SVM (RBF,C=100) | 93.1 | 92.5 | 89.8 | 93.8 | 91.2 | **92.0** |
| LBP+FPLBP+Dropout 0.5 (Best in [EEH14]) | - | - | - | - | - | 77.8 |

Tab. 2: Test results of gender estimators in five folds with mean (as accuracy percentage)

## 5   Conclusions

In this study, for automatic estimation of age and gender, which are both widely used soft biometric traits, transferring state-of-the-art deep Convolutional Neural Network models is explored. A generic object recognition model and a domain specific face recognition model are chosen to investigate the transferability of these models for age and gender classification. It is observed that i) both generic and domain specific deep CNN models can be successfully transferred for age and gender classification tasks, ii) transferring a deep CNN model can have better classification performance than training a task specific model from scratch in case of availability of limited data, and iii) transfering from a closer domain is more useful than transfering from a generic model. These observations are obtained by transferring generic AlexNet-like and domain specific VGG-Face CNN models and comparing them with task specific GilNet CNN model on challenging Adience benchmark. Experimental results show that transferred models outperform the state-of-the-art GilNet model both for age and gender classification tasks by 7% and 4.5%, respectively.

## Acknowledgement

This work was supported by TUBITAK project no. 113E067 and by a Marie Curie FP7 Integration Grant within the 7th EU Framework Programme.

## References

[ABD16]   Antipov, G.; Berrani, S.A.; Dugelay, J.L.: Minimalistic CNN-based ensemble model for gender prediction from face images. In: Pattern Recognition Letters. S. 59-65, 2016.

[BR07]   Baluja, S.; Rowley, H.A.: Boosting sex identification performance. In: IJCV. S. 111-119, 2007.

[Ch11]   Choi, S.E.; et.al.: Age estimation using a hierarchical classifier based on global and

local facial features. In: Pattern Recognition. S. 1262-1281, 2011.

[DBD14]    Danisman, T.; Bilasco, I.M.; Djeraba, C.: Cross-database evaluation of normalized raw pixels for gender recognition under unconstrained settings. In: ICPR. IEEE, S. 3144-3149, 2014.

[DER15]    Dantcheva, A.; Elia, P.; Ross, A.: What else does your biometrics data reveal? A survey on soft biometrics. In: Trans. on Inf. Forensics and Security. IEEE, S. 441-467, 2015.

[De12]    Demirkus, M. et.al.: Soft biometric trait classification from real-world face videos conditioned on head pose estimation. In: CVPR Workshops. IEEE, S. 130-137, 2012.

[EEH14]    Eidinger, E.; Enbar, R.; Hassner, T: Age and gender estimation of unfiltered faces. In: Trans. on Inf. Forensics and Security. IEEE, S. 2170–2179, 2014.

[Es15]    Escalera, S; et.al.: Chalearn 2015 apparent age and cultural event recognition: datasets and results. In: Proc. of ICCV Workshops. IEEE, 2015.

[FH08]    Fu, Y; Huang, T.S.: Human age estimation with regression on discriminative aging manifold. In: Trans. Multimedia. IEEE, S. 578-584, 2008.

[GA09]    Gao, F.; Ai, H.: Face age classification on consumer images with gabor feature and fuzzy lda method. In: Advances in Biometrics. Springer, S. 132-141, 2009.

[GC09]    Gallagher, A.C.; Chen, T.: Understanding images of groups of people. In: Proc. of CVPR. IEEE, S. 256-263, 2009.

[Gi14]    Girshick, R.; et.al: Rich feature hierarchies for accurate object detection and semantic segmentation. In: CVPR. IEEE, 2014.

[Gu08]    Guo, G.; et.al.: Locally adjusted robust regression for human age estimation. In: Workshop on Applications of Computer Vision. IEEE, S. 721-724, 2008.

[HL06]    Huang, F. J.; LeCun, Y.: Large-scale learning with svm and convolutional for generic object categorization. In: CVPR. IEEE, S. 284-291, 2006.

[Hu07]    Huang, G.B.; et.al.: Labeled faces in the wild: A database for studying face recognition in unconstrained environments. University of Massachusetts, Amherst. S. 7-49, 2007.

[JDN04a]    Jain, A.K.; Dass, S.C.; Nandakumar, K.: Can soft biometric traits assist user recognition? In (Jain, A.K.; Ratha, N.K.): Biometric Technology for Human Identification. SPIE, S. 561-572, 2004.

[JDN04b]    Jain, A.K.; Dass, S.C.; Nandakumar, K.: Soft biometric traits for personal recognition systems. In: Int. Conf. on Biometric Authentication. LNCS, S. 731-738, 2004.

[Ji14]    Jia, Y.; et.al.: Caffe: Convolutional architecture for fast feature embedding. In: Int. Conf. on Multimedia. ACM, S. 675-678, 2014.

[KCW15]    Kuang, Z.; Chen, H.; Wei, Z.: Deeply learned rich coding for cross-dataset facial age estimation. In: ICCV Workshops. IEEE, 2015.

[KL94]    Kwon, Y.H.; Lobo, N da Vitoria: Age classification from facial images. In: Proc. of CVPR. IEEE, S. 762-767, 1994.

[KSH12]   Krizhevsky, A.; Sutskever, I.; Hinton, G.E.: ImageNet classification with deep convolutional neural networks. In: NIPS, S. 1097-1105, 2012.

[Ku09]    Kumar, N., et.al.: Attribute and simile classifiers for face verification. In: Proc. of ICCV. IEEE, S. 365-372, 2009.

[La15]    Lanitis, A.: The FG-NET Aging Database, http://www-prima.inrialpes.fr/FGnet/html/benchmarks.html, Stand: 2015.

[LH15]    Levi, G.; Hassner, T.: Age and gender classification using convolutional neural networks. In: CVPR Workshops on Analysis and Modeling of Faces and Gestures. IEEE, S. 34-42, 2015.

[LTC02]   Lanitis, A.; Taylor, C.; Cootes, T.: Toward automatic simulation of aging effects on face images. In: Transactions on Pattern Analysis and Machine Intelligence. IEEE, S. 442-455, 2002.

[MAP16]   Mansanet, J.; Albiol, A.; Paredes, R.: Local deep neural networks for gender recognition. In: Pattern Recognition Letters. S. 80-86, 2016.

[MY02]    Moghaddam, B.; Yang, M.H.: Learning gender with support faces. In: Transactions on Pattern Analysis and Machine Intelligence. IEEE, S. 707–711, 2002.

[NPJ10]   Niinuma, K.; Park, U.; Jain, A.K.: Soft biometric traits for continuous user authentication. In: Trans. on Inf. Forensics and Security. IEEE, S. 771-780, 2010.

[Oq14]    Oquab, M.; et.al: Learning and transferring mid-level image representations using convolutional neural networks. In: CVPR. IEEE, 2014.

[Ph98]    Phillips, P.J., et.al.: The feret database and evaluation procedure for face-recognition algorithms. In: Image and vision computing. S. 295-306, 1998.

[Po15]    Pontes, Jhony K.; et.al.: A flexible hierarchical approach for facial age estimation based on multiple features. In: Pattern Recognition. 2015.

[PVZ15]   Parkhi, Omkar M.;Vedaldi, A.;Zisserman, A.: Deep face recognition. In:BMVC. 2015.

[Ra14]    Razavian, A.S., et.al.: CNN features off-the-shelf: an astounding baseline for recognition. In: CVPR Workshops. IEEE, S. 512-519, 2014.

[Ra15]    Ranjan, R.; et.al.: Unconstrained age estimation with deep convolutional neural networks. In: Proc. of ICCV Workshops. IEEE, S. 109-117, 2015.

[RTG15]   Rothe, R.; Timofte, R.; Van Gool, L.: DEX: Deep EXpectation of apparent age from a single image. In: Proc. of ICCV Workshops. IEEE, S. 10-15, 2015.

[RT06]    Ricanek, K.; Tesafaye, T.: MORPH: A longitudinal image database of normal adult age progression. In: Proc. 7th Int. Conf. on Automatic Face Gesture Recognition, S. 341–345, 2006.

[Ru15]    Russakovsky, O.; et.al.: Imagenet large scale visual recognition challenge. In: IJCV. 2015.

[SZ15]    Simonyan, K.; Zisserman, A.: Very deep convolutional networks for large-scale image recognition. In: ICLR. 2015.

[Yo14]    Yosinski, J.; et.al.: How transferable are features in deep neural networks? In: Advances in Neural Information Processing Systems. 2014.